%% file: divergenceOfTD1ijcnn.tex
\documentclass[journal]{IEEEtran}
\ifCLASSINFOpdf
   \usepackage[pdftex]{graphicx}
\else
   \usepackage[dvips]{graphicx}
\fi% graphicx was written by David Carlisle and Sebastian Rahtz. It is
 \usepackage{amsmath} 
 \usepackage{amsfonts} 
 \usepackage{color}
\newcommand{\fracpartial}[2]{\frac{\partial {#1}}{\partial {#2}}}
\newcommand{\fpevalat}[3]{\left(\fracpartial{#1}{#2}\right)_{#3}}
\newcommand{\bigDevalat}[3]{\left(\bigD{#1}{#2}\right)_{#3}}
\newcommand{\bigD}[2]{\frac{D {#1} }{ D {#2}}}

\newcommand{\Gtarget}{{G'}}
\newcommand{\Vtarget}{{R^{\lambda}}}
\newcommand{\Vapprox}{{\widetilde{V}}}
\newcommand{\Gapprox}{{\widetilde{G}}}
\newcommand{\Qapprox}{{\widetilde{Q}}}
\newcommand{\Qtarget}{{{Q^{\lambda}}}}
\newcommand{\vecx}{\vec{x}}
\newcommand{\veca}{\vec{a}}
\newcommand{\reward}{r}
\newcommand{\model}{{f}}
\newcommand{\action}{{\pi}}

\newcommand{\state}{x}
\newcommand{\actiona}{a}
\newcommand{\StateSpace}{\mathbb{S}}
\newcommand{\ActionSpace}{\mathbb{A}}
\newcommand{\Weights}{\vec{w}}

\newcommand{\vecp}{\vec{p}}

\newcommand{\ddQdaSqared}{\frac{\partial {^2\Qapprox}}{\partial \veca\partial \veca}}
\newcommand{\ddQdaSqaredAt}[1]{\left(\ddQdaSqared\right)_{#1}}
\newcommand{\ifdf}[1]{#1}
\newcommand{\df}[0]{\ifdf{\gamma}}
\newcommand{\dfa}[1]{\ifdf{\gamma{#1}}}

\begin{document}
\setcounter{page}{3070} 
\title{The Divergence of Reinforcement Learning Algorithms with Value-Iteration and Function Approximation}

\author{Michael~Fairbank,~\IEEEmembership{Student Member,~IEEE} and Eduardo~Alonso
\thanks{M. Fairbank and E. Alonso are with the Department of Computing,
School of Informatics,
City University London, 
London, UK (e-mail: michael.fairbank.1@city.ac.uk; E.Alonso@city.ac.uk).}% <-this % stops a space
}
%\markboth{}%
%{}
\IEEEaftertitletext{\begin{small}Cite as: Michael Fairbank and Eduardo Alonso, {\it The Divergence of Reinforcement Learning Algorithms with Value-Iteration and Function Approximation}, In Proceedings of the IEEE International Joint Conference on Neural Networks, June 2012, Brisbane (IEEE IJCNN 2012), pp. 3070--3077. 

{\bf Errata:} See footnote \ref{footnote:errata1}\end{small}}

\maketitle

\begin{abstract}
%\boldmath
This paper gives specific divergence examples of value-iteration for several major Reinforcement Learning and Adaptive Dynamic Programming algorithms, when using a function approximator for the value function.  These divergence examples differ from previous divergence examples in the literature, in that they are applicable for a greedy policy, i.e. in a ``value iteration'' scenario.  Perhaps surprisingly, with a greedy policy, it is also possible to get divergence for the algorithms TD(1) and Sarsa(1).  In addition to these divergences, we also achieve divergence for the Adaptive Dynamic Programming algorithms HDP, DHP and GDHP.
\end{abstract}

{\keywords Adaptive Dynamic Programming, Reinforcement Learning, Greedy Policy, Value Iteration, Divergence}

% \fi
\IEEEpeerreviewmaketitle

\section{Introduction} \label{sec:introduction}

{A}{daptive} Dynamic Programming (ADP) \cite{adpAnIntroduction}  and Reinforcement Learning (RL) \cite{suttonbarto-1998} are similar fields of study that aim to make an agent learn actions that maximise a long-term reward function.  These algorithms often rely on learning a ``value function'' that is defined in Bellman's Principle of Optimality \cite{bellman57}.  When an algorithm attempts to learn this value function by a general smooth function approximator, while the agent is being controlled by a ``greedy policy'' on that approximated value function, then ensuring convergence of the learning algorithm is difficult.

It has so far been an open question as to whether divergence can occur under these conditions and for which algorithms.  In this paper we present a simple artificial test problem which we use to make many RL and ADP algorithms diverge with a greedy policy.  The value function learning algorithms that we consider are Sarsa($\lambda$) \cite{rummery94line}, TD($\lambda$) \cite{sutton88learning}, and the ADP algorithms Heuristic Dynamic Programming (HDP), Dual Heuristic Dynamic Programming (DHP), Globalized Dual Heuristic Dynamic Programming (GDHP) \cite{hic92ch13,prokhorovWunschACD,HLADPch3Ferarri} and Value-Gradient Learning (VGL($\lambda$)) \cite{fairbankAlonso11lorvgrtpgl,fairbankAlonso2012IJCNN_vgl}.  We prove divergence of all of these algorithms (including VGL(0), VGL(1), Sarsa(0), Sarsa(1), TD(0), TD(1), DHP and GDHP), all when operating with greedy policies, i.e. in a ``value-iteration'' setting.

Some of these algorithms have convergence proofs when a {\it fixed} policy is used.  For example TD($\lambda$) is proven to converge when $\lambda=1$ since it is then (and only then) true gradient descent on an error function \cite{sutton88learning}.  Also for $0 \leq \lambda \leq 1$, it is proven to converge by \cite{tsitsiklis96analysis} when the approximate value function is linear in its weight vector and learning is ``on-policy''.  Recent advancements in the RL literature have extended convergence conditions of variants of TD($\lambda$) to an ``off-policy'' setting \cite{Sutton09TDC}, and with non-linear function approximation of the value function \cite{maei09nonlinear}. However, all these proofs apply to a fixed policy instead of the greedy policy situation we consider here.

\cite{HLADPch3Ferarri} show that ADP processes will converge to optimal behaviour if the value function could be perfectly learned over all of state space at each iteration.  However in reality we must work with a function approximator for the value function with finite capabilities, so this assumption is not valid.  Working with a general quadratic function approximator, \cite{werbos98adap-org} proves the general instability of DHP and GDHP.  This analysis was for a fixed policy, so with a greedy policy convergence would presumably seem even less likely.  This paper confirms this.

A key insight into the difficulty of understanding convergence with a greedy policy is shown by lemma 7 of \cite{fairbankAlonso11lorvgrtpgl} that the dependency of a greedy action on the approximated value function is {\it primarily through the value-gradient}, i.e. the gradient of the value function with respect to the state vector.  We use a value-gradient analysis in this paper to understand the divergence of {\it all} of the algorithms being tested.  \cite{fairbankAlonso11lorvgrtpgl} and \cite{fairbank08} recently defined a value-function learning algorithm that is proven to converge under certain smoothness conditions, using a greedy policy and an arbitrary smooth approximated value function, so this contrasts greatly to the diverging algorithm examples we give here.

In the rest of this introduction (sections \ref{sec:definitions} to \ref{sec:endOfIntroduction}), we state the general RL/ADP problem and give the necessary function definitions.  In section \ref{sec:algorithms} we give definitions of the algorithms that we are testing.

The approach we make to achieve divergence is to define a problem that is simple enough to analyse algebraically, but flexible enough to provide a divergence example (sections \ref{sec:toyproblem} to \ref{sec:actorDefinition}).  We then analyse a trajectory for this problem (sections \ref{sec:trajAnalysis} to \ref{sec:GdashAnalysis}), so that we can write the VGL($\lambda$) weight update as a single dynamic system and hence examine what choice of parameters could be made to force this dynamic system to diverge (section \ref{sec:vglDivergence}).   The VGL($\lambda$) weight update is easier to analyse than the TD($\lambda$) one, since as mentioned above the greedy policy depends on the value-gradient, so in section \ref{sec:divergenceTD} we just use the same learning parameters that caused divergence for VGL($\lambda$) and find empirically that they cause the other algorithms to diverge too. 

Finally, in section \ref{sec:conclusions}, we discuss the difficulty of ensuring value-iteration convergence but its potential advantages compared to policy-iteration.

\subsection{RL and ADP Problem Definition and Notation} \label{sec:definitions}
The typical RL/ADP scenario is an agent wandering around in an environment (with state space $\StateSpace \subset \Re^n$), such that at time $t$ it has state vector $\vecx_t\in \StateSpace$. At each time $t$ the agent chooses an action $\veca_t$ (from an action space $\ActionSpace\subset \Re^n$) which takes it to the next state according to the environment's model function $\vecx_{t+1} = \model(\vecx_t,\veca_t)$, and gives it an immediate reward, $\reward_t$, given by the function $\reward_t = \reward(\vecx_t, \veca_t)$. In general these model functions $\model$ and $\reward$ can be stochastic functions. The agent keeps moving, forming a trajectory of states $(\vecx_0, \vecx_1, \ldots)$, which terminates if and when a designated terminal state is reached.  In RL/ADP, we aim to find a {\it policy} function, $\action(\vecx)$, that calculates which action $\veca=\action(\vecx)$ to take for any given state $\vecx$.  The objective of RL/ADP is to find a policy such that the expectation of the total discounted reward, $\left<\sum_{t} \dfa{^t} \reward_t \right>$, is maximised for any trajectory.  Here $\df\in [0,1]$ is a constant {\it discount factor} that specifies the importance of long term rewards over short term ones.

\subsection{Approximate Value Function (Critic) and its Gradient}

We define $\Vapprox(\vecx,\Weights)$ to be the real-valued scalar output of a smooth function approximator with weight vector $\Weights$ and input vector $\vecx$.  This is the ``approximate value function'', or ``critic''.   We define $\Gapprox(\vecx,\Weights)$ as the ``approximate value gradient'', or ``critic gradient'', to be $\Gapprox(\vecx,\Weights) \equiv \fracpartial{\Vapprox(\vecx,\Weights)}{\vecx}$.

Here and throughout this paper, a convention is used that all defined vector quantities are columns, whether they are coordinates, or derivatives with respect to coordinates.  So, for example, $\Gapprox$, $\fracpartial{\Vapprox}{\vecx}$ and  $\fracpartial{\Vapprox}{\Weights}$ are all columns.

\subsection{Greedy Policy}

The greedy policy is the function that always chooses actions as follows:
\begin{equation} \veca= \arg \max_{\veca \in \ActionSpace} (\Qapprox (\vecx, \veca, \Weights))  \ \ \ \forall \vecx \label{eqn:actorObjective} \end{equation}
where we define the approximate Q Value function as 
\begin{equation} \Qapprox (\vecx,\veca,\Weights)=\reward(\vecx,\veca)+\df \Vapprox(\model(\vecx,\veca), \Weights) \label{eqn:Qapprox} \end{equation}

\subsection{Actor-critic architectures} \label{sec:actorCritic}

If a non-greedy policy is used, then a separate policy function would be used. This could be represented by a second function approximator, known as the actor (the first function approximator being the critic).  The actor and the critic together are known as an actor-critic architecture.

Training of the actor and critic would take place iteratively and in alternating phases.  Policy iteration is the situation where the critic is trained to completion in between every actor update.  Value iteration is the situation where the actor is trained to completion in between each critic update.  

The intention of the actor's training weight update is to make the actor behave more like a greedy policy.  Hence value iteration is very much like using a greedy policy, since the {\it objective} of training an actor to completion is to make the actor behave just like a greedy policy.  Hence the divergence results we derive in this paper for a greedy policy are applicable to an actor-critic architecture with value-iteration, assuming the function approximator of the actor has sufficient flexibility to learn the greedy policy accurately enough (which is true for the actor we define in section \ref{sec:actorDefinition}).

\subsection{Trajectory Shorthand Notation}
Throughout this paper, all subscripted indices are what we call trajectory shorthand notation.  These refer to the time step of a trajectory and provide corresponding arguments $\vecx_t$ and $\veca_t$ where appropriate; so that for example $\Vapprox_{t+1}\equiv \Vapprox(\vecx_{t+1}, \Weights)$; $\Qapprox_{t+1}\equiv \Qapprox(\vecx_{t+1}, \veca_{t+1}, \Weights)$; $\fpevalat{\Qapprox}{\veca}{t}$ is shorthand for $\left.\fracpartial{\Qapprox(\vecx,\veca,\Weights)}{\veca}\right|_{(\vecx_t, \veca_t, \Weights)}$ and $\fpevalat{\Vapprox}{\Weights}{t}$ is shorthand for $\left.\fracpartial{\Vapprox(\vecx,\Weights)}{\Weights}\right|_{(\vecx_t, \Weights)}$.

\label{sec:endOfIntroduction}

\section{Learning Algorithms and Definitions} \label{sec:algorithms}

\subsection{TD($\lambda$) Learning}
\label{sec:tdlearning}
The TD($\lambda$) algorithm \cite{sutton88learning} can be defined in batch mode by the following weight update applied to an entire trajectory:
\begin{equation} \Delta \Weights = \alpha \sum_{t} \fpevalat{\Vapprox}{\Weights}{t}(\Vtarget_t-\Vapprox_t) \label{eqn:tdWeightUpdate} \end{equation}
where $\lambda \in [0,1]$, and $\alpha>0$ are fixed constants.  $\Vtarget$ is the (moving) target for this weight update.  It is known as the ``$\lambda$-Return'', as defined by \cite{watkins89}.  For a given trajectory, this can be written concisely using trajectory shorthand notation by the recursion
\begin{equation}
\Vtarget_t=r_t+\df (\lambda \Vtarget_{t+1}+(1-\lambda)\Vapprox_{t+1}) \label{eqn:VtargetDefinition}
\end{equation}
with $\Vtarget_t=0$ at any terminal state, as proven in Appendix A of \cite{fairbankAlonso11lorvgrtpgl}.  This equation introduces the dependency on $\lambda$ into eq. \ref{eqn:tdWeightUpdate}.  Using the $\lambda$-Return enables us to write TD($\lambda$) in this very concise way, known as the ``forwards view of TD($\lambda$)'' by \cite{suttonbarto-1998}, however the traditional way to implement the algorithm is using ``eligibility traces'', as described by \cite{sutton88learning}.

TD($\lambda$) is defined for the task of {\it policy evaluation}, i.e. it is defined just for the task of learning the approximated value function for a {\it fixed} policy.  It is not usually used with a greedy policy, which is the circumstance in which  we consider it in this paper.  However we show in section \ref{sec:sarsaDivergence} that the TD($\lambda$) weight update can be equivalent in some circumstances to the Sarsa($\lambda$) weight update, which {\it is} defined for a greedy policy.  Another reason to consider TD($\lambda$) with a greedy policy is that TD($\lambda$) can be used in an actor-critic architecture as part of a value-iteration scheme, which, as we described in section \ref{sec:actorCritic}, is very similar to using a greedy policy.  

\subsection {Sarsa($\lambda$) Algorithm} \label{sec:sarsaAppendix}
Sarsa($\lambda$) is an algorithm for control problems that learns to approximate the $\Qapprox(\vecx,\veca,\Weights)$ function \cite{rummery94line}.  It is designed for policies that are dependent on the $\Qapprox(\vecx,\veca,\Weights)$ function (e.g.  the greedy policy or a  greedy policy with added stochastic noise), where $\Qapprox(\vecx,\veca,\Weights)$ here is defined to be the output of a given function approximator.

The Sarsa($\lambda$) algorithm is defined for trajectories where all actions after the first are found by the given policy; the first action $\veca_0$ can be arbitrary.  The function-approximator update is defined to be:
\begin{equation} \Delta \Weights = \alpha \sum_{t} \fpevalat{\Qapprox}{\Weights}{t}(\Qtarget_t -\Qapprox_t) \label{eqn:sarsaWeightUpdate} \end{equation}
\noindent where $\Qtarget$ is the target for this weight update.  This is analogous to the $\lambda$-return, but uses the function approximator $\Qapprox$ in place of $\Vapprox$.  We can define $\Qtarget$ recursively in trajectory shorthand notation by
\begin{equation}
\Qtarget_t=r_t+\df (\lambda \Qtarget_{t+1}+(1-\lambda)\Qapprox_{t+1}) \label{eqn:QtargetDefinition} 
\end{equation}
with $\Qtarget_t=0$ at any terminal state.

\subsection{The VGL($\lambda$) Algorithm} 

To define the VGL($\lambda$) algorithm, throughout this paper we use a convention that differentiating a column vector function by a column vector causes the vector in the numerator to become transposed (becoming a row).  For example $\fracpartial{\model}{\vecx}$ is a matrix with element $(i,j)$ equal to $\fracpartial{\model(\vecx, \veca)^j}{\vecx^i}$.  Similarly, $\left(\fracpartial{\Gapprox}{\Weights}\right)^{ij}=\fracpartial{\Gapprox^j}{\Weights^i}$, and $\fpevalat{\Gapprox}{\Weights}{t}$ is this matrix evaluated at $(\vecx_t, \Weights)$.

Using this notation and the implied matrix products, all VGL algorithms can be defined by a weight update of the form:
\begin{equation} \Delta \Weights = \alpha \sum_{t} \fpevalat{\Gapprox}{\Weights}{t} \Omega_t (\Gtarget_t-\Gapprox_t) \label{eqn:NonResidGrads} \end{equation}
where $\alpha$ is a small positive constant;  $\Gapprox_t$ is the approximate value gradient; and $\Gtarget_t$ is the ``target value gradient'' defined recursively by:
\begin{small}
\begin{align}
\Gtarget_t =&\bigDevalat{\reward}{\vecx}{t}
+ \df \bigDevalat{\model}{\vecx}{t} 
	 \left(\lambda \Gtarget_{t+1}+(1-\lambda)\Gapprox_{t+1}\right) & \label{eqn:Gtarget} \end{align}
\end{small}
with $\Gtarget_t=\vec{0}$ at any terminal state; where $\Omega_t$ is an arbitrary positive definite matrix of dimension $(\dim{\vecx} \times \dim{\vecx})$; and where $\bigD{}{\vecx}$ is shorthand for
\begin{equation}
\bigD{}{\vecx}\equiv \fracpartial{}{\vecx}+\fracpartial{\action}{\vecx}\fracpartial{}{\veca} \ ;\label{eqn:bigDNotation}
\end{equation}
and where all of these derivatives are assumed to exist.   Equations \ref{eqn:NonResidGrads}, \ref{eqn:Gtarget} and \ref{eqn:bigDNotation}  define the VGL($\lambda$) algorithm.  \cite{fairbankAlonso11lorvgrtpgl} and \cite{fairbankAlonso2012IJCNN_vgl} give further details, and pseudocode for both on-line and batch-mode implementations.

The $\Omega_t$ matrix was  introduced by Werbos for the algorithm GDHP (e.g. see \cite[eq. 32]{werbos98adap-org}), and can be chosen freely by the experimenter, but it is in general difficult to decide how to do this; so for most purposes it is just taken to be the identity matrix.  However for the special choice of 
\begin{equation}
\Omega _t=\begin{cases}
-\fpevalat{f}{\veca}{t-1}^T
\ddQdaSqaredAt{t-1}^{-1}
\fpevalat{f}{\veca}{t-1} & \text{for $t>0$}\\
0 & \text{for $t=0$}\end{cases}
\label{eqn:Omega}, \end{equation}
the algorithm VGL(1) is proven to converge \cite{fairbankAlonso11lorvgrtpgl} when used in conjunction with a greedy policy, and under certain smoothness assumptions.

\subsection{Definition of the ADP Algorithms HDP, DHP and GDHP}

All of the ADP algorithms we will define here are particularly intended for the situation of actor-critic architectures.    However for our divergence examples in this paper we are instead using the greedy policy.  As detailed in section \ref{sec:actorCritic}, using an actor-critic architecture with value-iteration is very similar to using a greedy policy.  

The three ADP algorithms we consider here can all be defined in terms of the algorithms defined so far in this paper.

\begin{itemize}

\item The algorithm Heuristic Dynamic Programming (HDP) uses the same weight update for its $\Vapprox$ function as TD(0).

\item The algorithm Dual Heuristic Dynamic Programming (DHP) uses the same weight update for its $\Gapprox$ function as VGL(0).  In DHP, the function $\Gapprox(\vecx, \Weights)$ is usually implemented as the output of a separate {\it vector} function approximator, but in this paper's divergence example we don't do this (instead we use $\Gapprox \equiv \fracpartial{\Vapprox}{\vecx}$).  

\item Globalized Dual Heuristic Programming (GDHP) uses a linear combination of a weight update by VGL(0) and one by TD(0).
\end{itemize}

These ADP algorithms are traditionally used with a neural network to represent the critic.  But this is not always necessarily the case; any differentiable structure will suffice \cite{prokhorovWunschACD}.  In this paper we make use of simple quadratic functions to represent the critic.

\section{Problem Definition For Divergence}
 \label{sec:toyproblem}

We define the simple RL problem domain and function approximator suitable for providing divergence examples for the algorithms being tested.  

First we define an environment with $\vecx \in \Re$ and $\veca \in \Re$, and model functions:
\begin{subequations} \label{eqn:toyProblemModelFunctions}
\begin{align}
\model(\state_t, t, \actiona_t)&=  \begin{cases} \state_t+\actiona_t & \text{if $t \in \{0,1\}$} \\
 			\state_t & \text{if \(t = 2\)} \end{cases} \label{eqn:toyProblemModelFunctionsF}\\
\reward(\state_t, t, \actiona_t)&=\begin{cases} 
			 -k{\actiona_t}^2 & \text{if $t \in \{0,1\}$} \\
			 -{\state_t}^2 & \text{if \(t = 2\)} \end{cases} \end{align} 
\end{subequations}
where $k>0$ is a constant.  Each trajectory is defined to terminate at time step $t=3$, so that exactly three rewards are received by the agent (rewards are given at timings as defined in section \ref{sec:definitions}, i.e. with the final reward $r_2$ being received on transitioning from $t=2$ to $t=3$).   In these model function definitions, action $\actiona_2$ has no effect, so the whole trajectory is parametrised by just $\state_0$, $\actiona_0$ and $\actiona_1$, and the total reward for this trajectory is $-k({\actiona_0}^2+{\actiona_1}^2)-(\state_0+\actiona_0+\actiona_1)^2$. These model functions are dependent on $t$, which is an abuse of notation we have adopted for brevity, but this could be legitimised by including $t$ into $\vecx$.

The divergence example we derive below considers a trajectory which starts at $\state_0=0$.  From this start point, the optimal actions are $a_0=a_1=0$.

\subsection{Critic Definition} \label{sec:criticDefinition}

A critic function is defined using a weight vector with just four weights, $\Weights=(w_1,w_2,w_3,w_4)^T$:
\begin{align}\Vapprox(\state_t, t,\Weights)=\begin{cases} 
			-c_1{\state_1}^2+w_1 \state_1+w_3 & \hbox{if $t=1$} \\
			-c_2{\state_2}^2+w_2 \state_2+w_4 & \hbox{if $t=2$} \\
 			0 & \hspace{-0.2in} \hbox{if $t \in \{0,3\}$} \end{cases} \label{eqn:de_critic}
\end{align}
\noindent where $c_1$ and $c_2$ are real positive constants.

Hence the critic gradient function, $\Gapprox \equiv \fracpartial{\Vapprox}{\state}$, is given by:
\begin{align}\Gapprox(\state_t, t, \Weights)= \begin{cases} 
			-2c_t{\state_t}+w_t& \hbox{if $t \in \{1,2\}$} \\
 			0 & \hbox{if $t \in \{0,3\}$} \end{cases} \label{eqn:GapproxDivergenceExample} \end{align}

We note that this implies 
\begin{align}\fpevalat{\Gapprox}{\Weights^k}{t}= \begin{cases}
                     1 & \text{if $t \in \{1,2\}$ and $t=k$} \\
                     0 & \text{otherwise}
                                                 \end{cases} \label{eqn:de_dGdw}
\end{align}
\subsection{Actor Definition} \label{sec:actorDefinition}
In this problem it is possible to define a function approximator for the actor with sufficient flexibility to behave exactly like a greedy policy, provided the actor is trained in a value-iteration scheme.  This is particularly easy to do here, since the trajectory is defined to have a fixed start point $\state_0=0$.  For example, if we define the weight vector of the actor,  $\vec{z}$, to have just two components, so that $\vec{z}=(z_0,z_1)$, and then define the output of the actor to be the identity function of these two weights, so that $a_0 \equiv z_0$ and $a_1\equiv z_1$, then training the actor to completion would be equivalent to solving the greedy policy's maximum condition.    This enables the divergence results of this paper to also apply to actor-critic architectures, as discussed in section \ref{sec:actorCritic}.

\subsection{Unrolling a greedy trajectory} \label{sec:trajAnalysis}

Substituting the model functions (eq. \ref{eqn:toyProblemModelFunctions}) and the critic definition (eq. \ref{eqn:de_critic}) into the $\Qapprox$ function definition (eq. \ref{eqn:Qapprox}) gives, with $\df=1$,
\begin{small}
\begin{align*}
&\Qapprox(\state_t, t, \actiona_t, \Weights)\\
&=\begin{cases}
-k(\actiona_0)^2-c_1(\state_0+\actiona_0)^2+w_1(\state_0+\actiona_0)+w_3 & \text{if $t=0$} \\
-k(\actiona_1)^2-c_2(\state_1+\actiona_1)^2+w_2(\state_1+\actiona_1)+w_4 & \text{if $t=1$} \\
\end{cases}
\end{align*}
\end{small}
In order to maximise this with respect to $\actiona_t$ and get greedy actions, we first differentiate to get,
\begin{small}
\begin{align}
\fpevalat{\Qapprox}{\actiona}{t}&=-2k\actiona_t-2c_{t+1}(\state_t+\actiona_t)+w_{t+1} &\text{for $t\in\{0,1\}$}\nonumber\\
&=-2\actiona_t(c_{t+1}+k)+w_{t+1}-2c_{t+1}\state_t &\text{for $t\in\{0,1\}$} \label{eqn:dQda}
\end{align}
\end{small}
Hence the greedy actions are given by
\begin{align}
\actiona_0 & \equiv \frac{w_1-2c_1\state_0}{2(c_1+k)} \label{eqn:de_greedyAct0} \\
\actiona_1 & \equiv \frac{w_2-2c_2\state_1}{2(c_2+k)} \label{eqn:de_greedyAct1} \end{align}  
Following these actions along a trajectory starting at $\state_0=0$, and using the recursion $\state_{t+1}=\model(\state_t, \actiona_t)$ with the model functions (eq. \ref{eqn:toyProblemModelFunctions}) gives \begin{align}                                                                                                                                              
\state_1&=\actiona_0=\frac{w_1}{2(c_1+k)} \label{eqn:de_x1}\\
\text{and }\state_2&=\state_1+\actiona_1=\frac{w_2 (c_1+k)+k w_1}{2(c_2+k)(c_1+k)}  \label{eqn:de_x2}                                                                                                                            \end{align}
Substituting $\state_1$ (eq. \ref{eqn:de_x1}) back into the equation for $\actiona_1$ (eq. \ref{eqn:de_greedyAct1}) gives $\actiona_1$ purely in terms of the weights and constants:\footnote{We emphasise that we are doing this step for the divergence analysis, and that this is {\it not} the way that VGL is meant to be implemented in practice.}
\begin{align}
\actiona_1 & \equiv \frac{w_2 (c_1+k)-c_2w_1}{2(c_2+k)(c_1+k)} \label{eqn:de_greedyAct1b} \end{align}

\subsection{Evaluation of value-gradients along the greedy trajectory}
We can now evaluate the $\Gapprox$ values by substituting the greedy trajectory's state vectors (eqs. \ref{eqn:de_x1}-\ref{eqn:de_x2}) into  eq. \ref{eqn:GapproxDivergenceExample}, giving:
\begin{align}
\Gapprox_1&=-\frac{c_1 w_1}{(c_1+k)}+w_1=\frac{w_1 k}{(c_1+k)}\label{eqn:de_Gapprox1}\\
\text{and }\Gapprox_2&=-\frac{w_2 (c_1+k)c_2+k w_1 c_2}{(c_2+k)(c_1+k)}+w_2 \nonumber\\
 &=\frac{w_2k(c_1+k)-k w_1 c_2}{(c_2+k)(c_1+k)} \label{eqn:de_Gapprox2}
\end{align}

The greedy actions in equations \ref{eqn:de_greedyAct0} and \ref{eqn:de_greedyAct1} both satisfy
\begin{align}
 \fpevalat{\action}{\state}{t}=\begin{cases}
  \frac{-c_{t+1}}{c_{t+1}+k} & \text{for $t \in \{0,1\}$} \\
  0		     & \text{otherwise} 
                              \end{cases} \label{eqn:dAdxt}
\end{align}

Substituting eqs. \ref{eqn:dAdxt} and \ref{eqn:toyProblemModelFunctions} into $\bigD{\model}{\state}=\fracpartial{\model}{\state}+\fracpartial{\action}{\state}\fracpartial{\model}{\actiona}$ gives 
\begin{small}
\begin{align}
 \bigDevalat{\model}{\state}{t}=\begin{cases}
1-\frac{c_{t+1}}{c_{t+1}+k}=\frac{k}{c_{t+1}+k} & \text{if $t \in \{0,1\}$} \\
1				    & \text{if $t=2$}                                
                               \end{cases} \label{eqn:de_dmodeldx}
\end{align}
\end{small}

Similarly, substituting them into $\bigD{\reward}{\state}=\fracpartial{\reward}{\state}+\fracpartial{\action}{\state}\fracpartial{\reward}{\actiona}$ gives
\begin{small}
\begin{align}
 \bigDevalat{\reward}{\state}{t}&=\begin{cases}
0-\frac{c_{t+1}}{c_{t+1}+k}(-2k\actiona_t)=\frac{2kc_{t+1}\actiona_t}{c_{t+1}+k} & \text{if $t \in \{0,1\}$} \\
-2\state_t				    & \text{if $t=2$}                                
                               \end{cases} \label{eqn:de_drewarddx}
\end{align}
\end{small}

\subsection{Backwards pass along trajectory} \label{sec:GdashAnalysis}
We do a backwards pass along the trajectory calculating the target gradients using eq. \ref{eqn:Gtarget} with $\df=1$, and starting with $\Gapprox_3=0$ (by eq. \ref{eqn:GapproxDivergenceExample}) and $\Gtarget_3=0$ (since $\Gtarget_3$ is at a terminal state):
\begin{small}
\begin{align}
 \Gtarget_2 =&\bigDevalat{\reward}{\state}{2} &\hspace{-0.2in}\text{by eq. \ref{eqn:Gtarget} and $\Gtarget_3=\Gapprox_3=0$}\nonumber \\
=&-2\state_2 \nonumber & \text{by eq. \ref{eqn:de_drewarddx}} \nonumber \\
=&-\frac{w_2 (c_1+k)+k w_1}{(c_2+k)(c_1+k)} \label{eqn:de_Gtarget2} & \text{by eq. \ref{eqn:de_x2}} 
\end{align}
\end{small}

Similarly, 
\begin{small}
\begin{align}
 \Gtarget_1 =&\bigDevalat{\reward}{\state}{1}+\bigDevalat{\model}{\state}{1}\left(\lambda\Gtarget_2+(1-\lambda)\Gapprox_2\right) &\text{by eq. \ref{eqn:Gtarget}}\nonumber \\
=&\frac{2kc_2\actiona_1}{c_2+k}+\frac{k}{c_2+k} \left(\lambda\Gtarget_2+(1-\lambda)\Gapprox_2\right) &\hspace{-0.2in}\text{by eqs. \ref{eqn:de_drewarddx},\ref{eqn:de_dmodeldx}}\nonumber \\
=&\frac{kc_2(w_2 (c_1+k)-c_2w_1)}{(c_1+k)(c_2+k)^2}
&\nonumber \\&
+\frac{k}{c_2+k} 
\left(-\lambda\frac{w_2 (c_1+k)+k w_1}{(c_2+k)(c_1+k)}
\right.&\nonumber \\&\left.
+(1-\lambda)\frac{w_2k(c_1+k)-k w_1 c_2}{(c_2+k)(c_1+k)}\right) &\hspace{-.2in}\text{by eqs.\ref{eqn:de_greedyAct1b},\ref{eqn:de_Gapprox2},\ref{eqn:de_Gtarget2}}\nonumber \\
=&\frac{w_2 k(c_2-\lambda+k(1-\lambda))}{(c_2+k)^2}
&\nonumber \\&
-\frac{w_1k(k\lambda +(c_2)^2+k(1-\lambda)  c_2)}{(c_1+k)(c_2+k)^2}
 & \label{eqn:de_Gtarget1}\end{align}
\end{small}

\section{Divergence Examples for VGL and DHP Algorithms} \label{sec:vglDivergence}

We now have the whole trajectory and the terms $\Gapprox$ and $\Gtarget$ written algebraically, so that we can next analyse the VGL($\lambda$) weight update for divergence.

The VGL($\lambda$) weight update (eq. \ref{eqn:NonResidGrads}) combined with $\Omega_t$=1 gives
\begin{subequations} 
\begin{align} \Delta w_i &= \alpha \sum_{t} \fpevalat{\Gapprox}{w_i}{t} (\Gtarget_t-\Gapprox_t) &\label{eqn:de_weightUpdateA} \\
&= \alpha   (\Gtarget_i-\Gapprox_i) & \hspace{-0.5in}\text{(for $i \in \{1,2\}$, by eq. \ref{eqn:de_dGdw})} \nonumber \\%\end{align}
%\begin{align}
\Rightarrow \begin{pmatrix}\Delta w_1 \\ \Delta w_2 \end{pmatrix}&=\alpha 
\begin{pmatrix}\Gtarget_1-\Gapprox_1 \\ \Gtarget_2-\Gapprox_2 \end{pmatrix} & \nonumber \\
%\hspace{-0.5in}\Rightarrow \sum_{t} \fpevalat{\Gapprox}{\Weights}{t} (\Gtarget_t-\Gapprox_t) &=
%\begin{pmatrix}\Gtarget_1-\Gapprox_1 \\ \Gtarget_2-\Gapprox_2 \end{pmatrix} & \nonumber \\
&=\alpha A \begin{pmatrix}w_1 \\ w_2 \end{pmatrix} &\label{eqn:de_weightUpdateB}
 \end{align}
\end{subequations} 
where $A$ is a $2 \times 2$ matrix with elements found by subtracting equations \ref{eqn:de_Gapprox1} and \ref{eqn:de_Gapprox2} from equations \ref{eqn:de_Gtarget1} and \ref{eqn:de_Gtarget2}, respectively, giving,
\begin{align}
A=&\begin{pmatrix}
-\frac{k(k\lambda +(c_2)^2+k(1-\lambda)  c_2)}{(c_1+k)(c_2+k)^2}-\frac{k}{(c_1+k)}&
\frac{k(c_2+k-\lambda(k+1))}{(c_2+k)^2}\\
\frac{k (c_2-1)}{(c_2+k)(c_1+k)} &
\frac{-1-k}{(c_2+k)}
\end{pmatrix} \label{eqn:Amatrix}
\end{align}

Equation \ref{eqn:de_weightUpdateB} is the VGL($\lambda$) weight update written as a single dynamic system of just {\it two} variables, i.e. a shortened weight vector,  $\Weights=(w_1, w_2)^T$.   For this shortened weight vector, $\Weights$, by looking at the right-hand sides of the sequence of equations from eq. \ref{eqn:de_weightUpdateA} to eq. \ref{eqn:de_weightUpdateB}, we can conclude that
\begin{align}
\sum_{t} \fpevalat{\Gapprox}{\Weights}{t} (\Gtarget_t-\Gapprox_t)&= A \Weights \label{eqn:de_weightUpdateC} 
\end{align}

To add further complexity to the system, in order to achieve the desired divergence, we next define these two weights to be a linear function of two {\it other} weights, $\vecp=(p_1,p_2)^T$, such that the shortened weight vector is given by $\Weights = F \vecp$, where $F$ is a $2 \times 2$ constant real matrix.  The VGL($\lambda$) weight update equation can now be recalculated for these new weights, as follows:
\begin{small}\begin{align} 
\Delta \vecp &= \alpha \sum_{t} \fpevalat{\Gapprox}{\vecp}{t} (\Gtarget_t-\Gapprox_t) &\text{by eq. \ref{eqn:NonResidGrads} and $\Omega_t$=1} \nonumber \\
&=\alpha \sum_{t} \fracpartial{\Weights}{\vec{p}} \fpevalat{\Gapprox}{\Weights}{t} (\Gtarget_t-\Gapprox_t) &\text{by chain rule} \nonumber \\
&=\alpha \fracpartial{\Weights}{\vec{p}}\sum_{t}  \fpevalat{\Gapprox}{\Weights}{t} (\Gtarget_t-\Gapprox_t) &\text{since  independent of $t$} \nonumber \\
 &=\alpha\fracpartial{\Weights}{\vec{p}} A \Weights &\text{by eq. \ref{eqn:de_weightUpdateC}} \nonumber \\
&=\alpha (F^T  A F)\vec{p}. &\hspace{-0.8in}\text{by $\Weights = F \vecp$ and $\fracpartial{\Weights}{\vec{p}}=\fracpartial{(F \vecp)}{\vec{p}}=F^T$ } 
\label{eqn:2stpDynamicSystemInP}
\end{align}
\end{small}

The optimal actions $\actiona_0=\actiona_1=0$ would be achieved by $\vec{p}=\vec{0}$.  To produce a divergence example, we want to ensure that $\vecp$ does {\it not} converge to $\vec{0}$.

Taking $\alpha>0$ to be sufficiently small, then the weight vector $\vecp$ evolves according to a continuous-time linear dynamic system given by eq. \ref{eqn:2stpDynamicSystemInP}, and this system is stable if and only if the matrix product $F^TAF$ is ``stable'' (i.e. if the real part of every eigenvalue of this matrix product is negative).

Choosing $\lambda=0$ and $c_1=c_2=k=0.01$ gives $A=\begin{pmatrix}-0.75&0.5\\-24.75&-50.5\end{pmatrix}$  (by equation \ref{eqn:Amatrix}).  Choosing $F=\begin{pmatrix}10& 1\\-1 & -1\end{pmatrix}$ makes $F^TAF=\begin{pmatrix}117.0&-38.25\\189.0&-27.0\end{pmatrix}$ which has eigenvalues $45\pm45.22i$.  Since the real parts of these eigenvalues are positive, eq. \ref{eqn:2stpDynamicSystemInP} will diverge for VGL(0) (i.e. DHP).  In an extended analysis, we found that these parameters also cause VGL(0) to diverge  when the $\Omega_t$ matrices are included according to equation \ref{eqn:Omega} (see the Appendix for further details).  

Since GDHP is a linear combination of DHP, which we have proven to diverge, and TD(0) (which we prove to diverge below), it follows that GDHP can diverge with a greedy policy too.

Also, perhaps surprisingly, it is possible to get instability with VGL(1).  Choosing $c_2=k=0.01$, $c_1=0.99$ gives $A=\begin{pmatrix}-0.2625&-24.75\\-0.495&-50.5\end{pmatrix}$. Choosing $F=\begin{pmatrix}-1&-1\\.2&.02\end{pmatrix}$ makes $F^TAF=\begin{pmatrix}2.7665& 0.1295\\4.4954& 0.2222\end{pmatrix}$ which has two positive real eigenvalues.  Therefore this VGL(1) system diverges.  

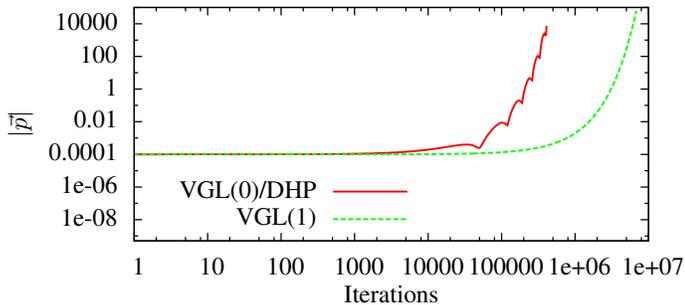
\begin{figure}[ht]
  \begin{center}
    \begin{tabular}{c}
        \hspace{0.2in}\resizebox{75mm}{!}{\input{plotVGL0and1divergencet}}
     \end{tabular}
 \caption {Divergence for VGL(0) (i.e. DHP) and VGL(1) using the learning parameters described in section \ref{sec:vglDivergence} and a learning rate of $\alpha=10^{-6}$.}
 \label{fig:divergingVGL0and1}
   \end{center}
 \end{figure}

Diverging weights are shown for the VGL(0) and VGL(1) algorithms in Figure \ref{fig:divergingVGL0and1}, with a learning rate of $\alpha=10^{-6}$.  Both experiments (and all subsequent experiments in this paper) used a starting weight vector of ($p_1,p_2,w_3,w_4)=(5.23*10^{-5}, 8.53*10^{-5},0, 0$), which is based upon a principal eigenvector of the $F^TAF$ matrix found to make VGL(1) diverge.

The divergence result for VGL(1) does not affect the convergence result by \cite{fairbankAlonso11lorvgrtpgl} which is for VGL(1) but with the special choice of $\Omega_t$ given by eq. \ref{eqn:Omega}, which we will refer to as VGL$\Omega$(1).  It was not possible to make VGL$\Omega$(1) diverge with the methods of this paper (see Appendix for futher details).  Figure \ref{fig:convergenceVGLOmega1} shows VGL$\Omega$(1) converging using the same learning parameters that made VGL(1) diverge.

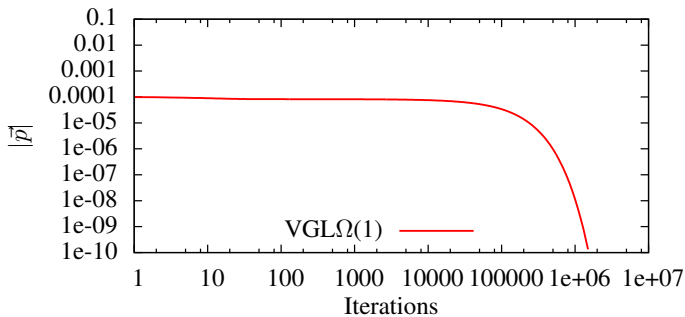
\begin{figure}[ht]
  \begin{center}
    \begin{tabular}{c}
         \hspace{0.2in}\resizebox{75mm}{!}{\input{plotVGLOmega1convergencet}}
     \end{tabular}
 \caption {Convergence for VGL$\Omega$(1) using the same parameters that caused VGL(1) to diverge, and $\alpha=10^{-3}$.  This algorithm demonstrates that it {\it is} possible to have proven convergence for a critic learning algorithm with a greedy policy and general function approximation.
}
 \label{fig:convergenceVGLOmega1}
   \end{center}
 \end{figure}

\section{Divergence results for TD($\lambda$), Sarsa($\lambda$) and HDP} \label{sec:divergenceTD}

To satisfy the requirement for exploration in TD($\lambda$)-based algorithms, we supplemented the greedy policies (eqs. \ref{eqn:de_greedyAct0} \& \ref{eqn:de_greedyAct1}) with a small amount of stochastic Gaussian noise with zero mean and variance 0.0001.  This Gaussian noise was necessary, since it is well known that these classic RL algorithms must be supplemented with some form of exploration.  This is the classic ``exploration versus exploitation'' dilemma.  Without exploration,  these algorithms do not converge to an optimal policy, in general.  Specific examples of converging to the wrong policy without exploration are given by \cite[sec. IV.H]{fairbankAlonso2012IJCNN_dhpTdComparison} and \cite[appendix B]{fairbank08}.

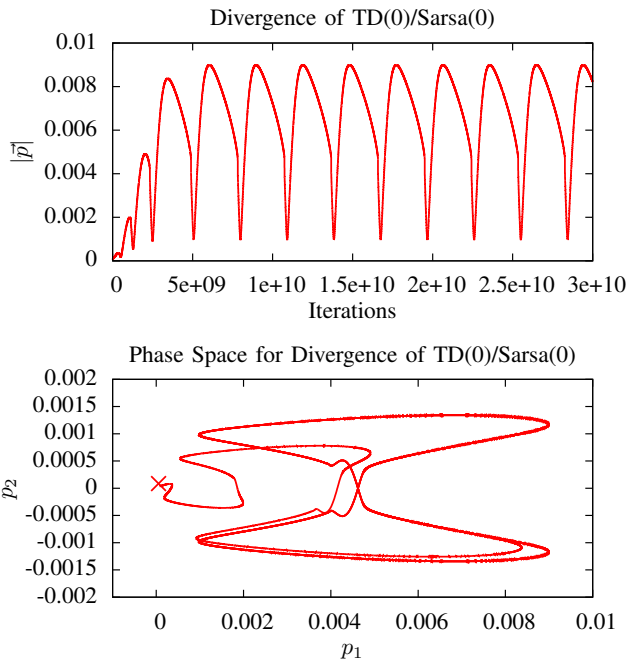
\begin{figure}[ht]
  \begin{center}
    \begin{tabular}{l}% \hspace{-6mm}
        \resizebox{70mm}{!}{\input{plotTD0divergencet}} %& \hspace{4mm} 
\\ \\
        \resizebox{70mm}{!}{\input{plotTD0divergencePhaseSpacet}}
    \end{tabular}
\caption {Divergence for TD(0) and Sarsa(0) generated with the diverging parameters described in section \ref{sec:divergenceTD}, and a learning rate of $\alpha=10^{-6}$. The upper graph shows progress of $|\vec{p}|$ versus iterations. The lower graph shows the corresponding evolution of the weight vector components $(p_1,p_2)$ in phase space.  This phase curve {\it starts} close to the origin (at the `X'), and finishes off in a limit cycle.}
\label{fig:divergenceTD0}
  \end{center}
\end{figure}

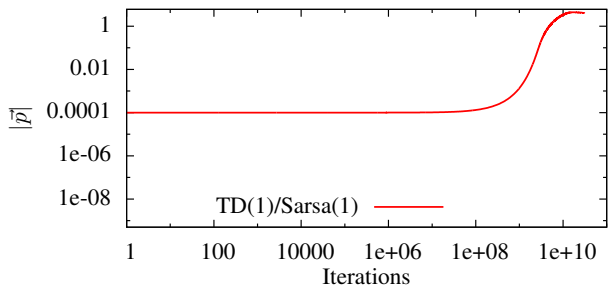
\begin{figure}[ht]
  \begin{center}
    \begin{tabular}{c}
        \resizebox{70mm}{!}{\input{plotTD1divergencet}}
    \end{tabular}
\caption {Divergence for TD(1) and Sarsa(1) generated with the diverging parameters described in section \ref{sec:divergenceTD}, and a learning rate of $\alpha=10^{-6}$.}
\label{fig:divergenceTD1}
  \end{center}
\end{figure}

To achieve divergence of these algorithms with the noisy greedy policy, we used exactly the same learning and environment constants as used for the VGL(0) and VGL(1) divergence experiments.  These choices of parameters, with the stochastic noise added to the greedy policy, made TD(0) and TD(1) diverge respectively, as shown in figures \ref{fig:divergenceTD0} and \ref{fig:divergenceTD1}.  %Source code for this is provided.  
Hence HDP diverges too, since this is equivalent to TD(0) with the given policy.

Although these divergence results for the TD($\lambda$) based algorithms were only found empirically, as opposed to the results for the previous sections which were first found analytically, these results do still have value.  Firstly, source code for the empirical experiments used here is provided by \cite[in ancillary files]{fairbankAlonso11divTD1}, so the empirical results should be entirely replicable.  Secondly, an insight into why the divergence parameters for VGL were sufficient to make the TD($\lambda$) based algorithms diverge too is because TD with stochastic exploration can be understood to be an approximation to a stochastic version of VGL($\lambda$), so we would {\it expect} a divergence example for VGL to cause divergence for TD($\lambda$) too.

Without the stochastic noise added to the greedy policy, these examples would not diverge, but instead converge to a sub-optimal policy, which is also considered a failure.

\subsection{Divergence results for Sarsa($\lambda$)} \label{sec:sarsaDivergence}
We next prove divergence for Sarsa($\lambda$) by choosing a function approximator for $\Qapprox$ that makes the Sarsa($\lambda$) weight update equivalent to the TD($\lambda$) weight update, so that the divergence result for TD($\lambda$) carries over to Sarsa($\lambda$).

 Sarsa($\lambda$) is designed to work with an arbitrary function approximator for $\Qapprox(\vecx, \veca, \Weights)$.  We will define our $\Qapprox$ function exactly by Eq. \ref{eqn:Qapprox}.  Rearranging eq. \ref{eqn:QtargetDefinition} gives
\begin{small}
\begin{align}
\left(\frac{\Qtarget_t-r_t}{\df}\right)
&=\lambda \Qtarget_{t+1}+(1-\lambda)\Qapprox_{t+1}&\nonumber\\
&=\lambda \Qtarget_{t+1}+(1-\lambda)(r_{t+1}+\df \Vapprox_{t+2})&\hspace{-0.5in}\text{by eq. \ref{eqn:Qapprox}}\nonumber\\
&=r_{t+1}+\lambda (\Qtarget_{t+1}-r_{t+1})+(1-\lambda)(\df \Vapprox_{t+2})&\nonumber\\
&=r_{t+1}+\df \left(\lambda \left(\frac{\Qtarget_{t+1}-r_{t+1}}{\df}\right)+(1-\lambda)\Vapprox_{t+2}\right)& \label{eqn:recursionQtargetMinusRt}
\end{align}
\end{small}

From this we can see that $\left(\frac{\Qtarget_{t}-r_{t}}{\df}\right)$ obeys the same recursion equation as $\Vtarget$, and they have the same endpoint (since both are zero at a terminal state), from which we can conclude (e.g. by comparing recursion equations \ref{eqn:recursionQtargetMinusRt} and \ref{eqn:VtargetDefinition}) that 
\begin{align*}
&\left(\frac{\Qtarget_{t}-r_{t}}{\df}\right) \equiv \Vtarget_{t+1} \\
\Rightarrow& \Qtarget_{t}=r_t+\df\Vtarget_{t+1}
\end{align*}
Substituting this into the Sarsa($\lambda$) weight update (eq. \ref{eqn:sarsaWeightUpdate}), with eq. \ref{eqn:Qapprox}, and simplifying gives
\begin{align*}
\Delta \Weights =&\alpha \sum_{t} \fpevalat{(\reward_t+\df \Vapprox(\vecx_{t+1}, \Weights))}{\Weights}{t}\left(
r_t 
\right.\\&\left.
+\df \Vtarget_{t+1}-(\reward_t+\df \Vapprox_{t+1})\right) \\
=& \alpha  \sum_{t} \df \fpevalat{\Vapprox}{\Weights}{t+1}\df(\Vtarget_{t+1}-\Vapprox_{t+1}) \\ 
=& \alpha \df^2 \sum_{t>0} \fpevalat{\Vapprox}{\Weights}{t}(\Vtarget_t-\Vapprox_t)\end{align*}
which is identical to TD($\lambda$) but with summation over $t$ now excluding $t=0$, and with an extra constant factor, $\df^2$.  The divergence example we derived above used $\df=1$, and had no weight update term for $t=0$, so uses an identical weight update.  Therefore this particular choice of function approximator for $\Qapprox$ and problem definition causes divergence for Sarsa($\lambda$) (with both $\lambda=1$ and $\lambda=0$).

\section{Conclusions} \label{sec:conclusions}

We have shown that under a value-iteration scheme, i.e. using a greedy policy, all of the RL algorithms have been made to diverge, and all but one of the VGL algorithms have been made to diverge.  The algorithm we found that didn't diverge was VGL$\Omega$(1) with $\Omega_t$ as defined by eq. \ref{eqn:Omega}, which is proven to converge by \cite{fairbankAlonso11lorvgrtpgl} and \cite{fairbank08} under these conditions.  %This value-iteration scheme can be implemented extremely computationally efficiently, if for the inner loop of finding the greedy policy, a closed form value-gradient policy as described by \cite{doya00reinforcement} is used.

These are new divergence results for TD(0), Sarsa(0), TD(1) and Sarsa(1), in that previous examples of divergence have only been for TD(0) and for non-greedy policies \cite{iii95residual,tsitsiklis96featurebased,tsitsiklis96analysis}.  The divergences we achieved for TD(1) and Sarsa(1) were only possible because of the use of a greedy policy (or equivalently, value-iteration).

A conclusion of this work is that the diverging algorithms considered cannot currently be reliably used for value-iteration, and instead can only be used under some form of policy iteration if provable convergence is required.  However there are some distinct advantages of value-iteration over policy-iteration.  Value-iteration using a greedy policy can be faster than using an actor-critic architecture.  Also policy iteration does provably converge in some cases \cite{suttonMcallester}, but the necessary conditions are thought to apply only when the function approximator for $\Vapprox$ is {\it linear} in the same features of the state vector that the function approximator for the policy uses as input (see footnote 1 of \cite{suttonMcallester}). 

The divergence results of this paper were derived for quadratic critic functions, as this was the situation that allowed for easiest analysis to derive concrete divergence examples.  We assume that similar divergence results will exist for neural network based critic functions, since neural networks are more complex structures that should allow for more possibilities for divergence situations similar to our simple example here.  In our experience, divergence often does occur when using a greedy policy with a neural network critic, but these situations are harder to analyse and make replicable.  In this situation, we speculate that a second order Taylor series expansion of the neural network could be made about the fixed point of the learning process, and locally this approximation could be behaving very similarly to the quadratic functions we have used in this paper.

It is hoped that the specific divergence examples of this paper will provide a better understanding of how value-iteration can diverge, and help motivate research to understand and prevent it.  We believe that the value-gradient analysis that produced the converging algorithm of Figure \ref{fig:convergenceVGLOmega1} by \cite{fairbankAlonso11lorvgrtpgl} could be helpful for reinforcement learning research, since this is a critic learning algorithm that does have convergence guarantees under a greedy policy with general function approximation.

%\vspace{1cm}
\appendix
In this appendix we give the extension analysis that was used to determine that VGL$\Omega(0)$ (i.e. VGL(0) with the $\Omega_t$ matrix of eq. \ref{eqn:Omega}) could be made to diverge.  We also include an analysis that shows VGL$\Omega(1)$ will converge in the experiment of this paper for any choice of experimental constants.

To construct the $\Omega_t$ matrix of eq. \ref{eqn:Omega}, first we note that differentiating equation \ref{eqn:toyProblemModelFunctionsF} gives
$\fpevalat{\model}{\actiona}{t}= 1$, for $t \in \{0,1\}$.  
And differentiating equation \ref{eqn:dQda} gives
\begin{align*}
\ddQdaSqaredAt{t}&=-2(c_{t+1}+k)&\text{for $t\in\{0,1\}$.}
\end{align*}
Hence, by equation \ref{eqn:Omega}, 
\begin{equation*}
\Omega _t=\begin{cases}
1/(2(c_{t}+k))
 & \text{for $t \in\{1,2\}$}\\
0 & \text{for $t=0$.}\end{cases}
\end{equation*}

The VGL($\lambda$) weight update can be re-derived using this new $\Omega_t$ matrix. Following the method that was used to derive equations \ref{eqn:de_weightUpdateA} to \ref{eqn:de_weightUpdateC}, but starting with this new $\Omega_t$ matrix, gives 
\begin{align*} \sum_{t} \fpevalat{\Gapprox}{\Weights}{t} \Omega_t (\Gtarget_t-\Gapprox_t)&=\alpha DA \begin{pmatrix}w_1 \\ w_2 \end{pmatrix} 
 \end{align*}
where $D=\begin{pmatrix}
\frac{1}{2(c_{1}+k)} &0 \\ 0 &\frac{1}{2(c_{2}+k)}        
\end{pmatrix}$ and $A$ is given by equation \ref{eqn:Amatrix}.  Then, defining $\Weights = F \vecp$ for a constant matrix $F$ (as done in section \ref{sec:vglDivergence}), and following the method that was used to derive eq. \ref{eqn:2stpDynamicSystemInP}, we would derive the VGL($\lambda$) weight update for the weight vector $\vecp$ as
\begin{align*} 
\Delta \vecp &= \alpha (F^T D A F)\vec{p}.
\end{align*}
As before, this system will converge for sufficiently small $\alpha$ if and only if the product $F^T D A F$ is ``stable'', i.e. if the real parts of the eigenvalues are negative. 

\subsection{Divergence of VGL$\Omega$(0)}
Choosing the same parameters that made VGL(0) diverge, i.e. $c_1=c_2=k=0.01$, gives $D=\begin{pmatrix}25&0\\0&25                                                                                                                                                                    \end{pmatrix}$.  Since $D$ is a positive multiple of the identity matrix, its presence will not affect the stability of the product $F^T D A F$, so the system for $\vecp$ will still be unstable, and diverge, just as it did for VGL(0).

\subsection{Convergence of VGL$\Omega$(1)}
When VGL$\Omega$(1) is used, convergence can be proven for any choice of parameters as follows:   When $\lambda=1$, the $A$ matrix of eq. \ref{eqn:Amatrix} reduces to\footnote{This version of this document contains a fix to the following equations - the constant factor 2 was missing from the version published in the proceedings of IJCNN12.\label{footnote:errata1}}
\begin{align*}
A=&\begin{pmatrix}
-\frac{k(k +(c_2)^2)}{(c_1+k)(c_2+k)^2}-\frac{k}{(c_1+k)}&
\frac{k(c_2-1)}{(c_2+k)^2}\\
\frac{k (c_2-1)}{(c_2+k)(c_1+k)} &
\frac{-1-k}{(c_2+k)}
\end{pmatrix} \\
=&2\begin{pmatrix}
-\frac{k(k +(c_2)^2)}{(c_2+k)^2}-k&
\frac{k(c_2-1)}{(c_2+k)}\\
\frac{k (c_2-1)}{(c_2+k)} &
-1-k
\end{pmatrix}D \\
=&2E\begin{pmatrix}
-{k(k +(c_2)^2+(c_2+k)^2)}&
{k(c_2-1)}\\
{k (c_2-1)} &
-1-k
\end{pmatrix}ED
\end{align*}
where $E=\begin{pmatrix}\frac{1}{(c_2+k)}&0\\0&1\end{pmatrix}$.  Hence the matrix  product $F^T D A F$ can now be written as 
$2F^TDEBEDF$ where $B=\begin{pmatrix}
-{k(k +(c_2)^2+(c_2+k)^2)}&
{k(c_2-1)}\\
{k (c_2-1)} &
-1-k
\end{pmatrix}$.  This new product is real and symmetrical (as we would expect it to be for true gradient descent), hence it has real eigenvalues.  %Note that this symmetry is a necessary condition for true gradient descent, and VGL$\Omega$(1) is proven to be gradient descent \cite{fairbankAlonso11lorvgrtpgl}. 
For any $c_2>0$ and $k>0$, the central matrix $B$ has a negative trace, and a determinant equal to $k(k+2)(k+c_2)^2$, which is positive.  Hence $B$ has two negative real eigenvalues.  Therefore, assuming $F$ is a full-rank matrix, the matrix product $2F^TDEBEDF$ must be negative definite, and therefore stable, and thus the dynamic system for $\vecp$ will converge.

%\vspace{1cm}



%\bibliographystyle{IEEEtran}
%\begin{thebibliography}{1}


%\bibliographystyle{IEEEtran}
%\bibliography{../../../MikeMasterBibtex}
\input{divergenceOfTD1ijcnn.bbl}

%\end{thebibliography}

\end{document}

%% file: plotVGL0and1divergencet.tex
% GNUPLOT: LaTeX picture with Postscript
\begingroup
  \makeatletter
  \providecommand\color[2][]{%
    \GenericError{(gnuplot) \space\space\space\@spaces}{%
      Package color not loaded in conjunction with
      terminal option `colourtext'%
    }{See the gnuplot documentation for explanation.%
    }{Either use 'blacktext' in gnuplot or load the package
      color.sty in LaTeX.}%
    \renewcommand\color[2][]{}%
  }%
  \providecommand\includegraphics[2][]{%
    \GenericError{(gnuplot) \space\space\space\@spaces}{%
      Package graphicx or graphics not loaded%
    }{See the gnuplot documentation for explanation.%
    }{The gnuplot epslatex terminal needs graphicx.sty or graphics.sty.}%
    \renewcommand\includegraphics[2][]{}%
  }%
  \providecommand\rotatebox[2]{#2}%
  \@ifundefined{ifGPcolor}{%
    \newif\ifGPcolor
    \GPcolortrue
  }{}%
  \@ifundefined{ifGPblacktext}{%
    \newif\ifGPblacktext
    \GPblacktextfalse
  }{}%
  % define a \g@addto@macro without @ in the name:
  \let\gplgaddtomacro\g@addto@macro
  % define empty templates for all commands taking text:
  \gdef\gplbacktext{}%
  \gdef\gplfronttext{}%
  \makeatother
  \ifGPblacktext
    % no textcolor at all
    \def\colorrgb#1{}%
    \def\colorgray#1{}%
  \else
    % gray or color?
    \ifGPcolor
      \def\colorrgb#1{\color[rgb]{#1}}%
      \def\colorgray#1{\color[gray]{#1}}%
      \expandafter\def\csname LTw\endcsname{\color{white}}%
      \expandafter\def\csname LTb\endcsname{\color{black}}%
      \expandafter\def\csname LTa\endcsname{\color{black}}%
      \expandafter\def\csname LT0\endcsname{\color[rgb]{1,0,0}}%
      \expandafter\def\csname LT1\endcsname{\color[rgb]{0,1,0}}%
      \expandafter\def\csname LT2\endcsname{\color[rgb]{0,0,1}}%
      \expandafter\def\csname LT3\endcsname{\color[rgb]{1,0,1}}%
      \expandafter\def\csname LT4\endcsname{\color[rgb]{0,1,1}}%
      \expandafter\def\csname LT5\endcsname{\color[rgb]{1,1,0}}%
      \expandafter\def\csname LT6\endcsname{\color[rgb]{0,0,0}}%
      \expandafter\def\csname LT7\endcsname{\color[rgb]{1,0.3,0}}%
      \expandafter\def\csname LT8\endcsname{\color[rgb]{0.5,0.5,0.5}}%
    \else
      % gray
      \def\colorrgb#1{\color{black}}%
      \def\colorgray#1{\color[gray]{#1}}%
      \expandafter\def\csname LTw\endcsname{\color{white}}%
      \expandafter\def\csname LTb\endcsname{\color{black}}%
      \expandafter\def\csname LTa\endcsname{\color{black}}%
      \expandafter\def\csname LT0\endcsname{\color{black}}%
      \expandafter\def\csname LT1\endcsname{\color{black}}%
      \expandafter\def\csname LT2\endcsname{\color{black}}%
      \expandafter\def\csname LT3\endcsname{\color{black}}%
      \expandafter\def\csname LT4\endcsname{\color{black}}%
      \expandafter\def\csname LT5\endcsname{\color{black}}%
      \expandafter\def\csname LT6\endcsname{\color{black}}%
      \expandafter\def\csname LT7\endcsname{\color{black}}%
      \expandafter\def\csname LT8\endcsname{\color{black}}%
    \fi
  \fi
  \setlength{\unitlength}{0.0500bp}%
  \begin{picture}(4492.80,2520.00)%
    \gplgaddtomacro\gplbacktext{%
      \csname LTb\endcsname%
      \put(132,609){\makebox(0,0)[r]{\strut{} 1e-08}}%
      \put(132,869){\makebox(0,0)[r]{\strut{} 1e-06}}%
      \put(132,1129){\makebox(0,0)[r]{\strut{} 0.0001}}%
      \put(132,1390){\makebox(0,0)[r]{\strut{} 0.01}}%
      \put(132,1650){\makebox(0,0)[r]{\strut{} 1}}%
      \put(132,1910){\makebox(0,0)[r]{\strut{} 100}}%
      \put(132,2170){\makebox(0,0)[r]{\strut{} 10000}}%
      \put(264,220){\makebox(0,0){\strut{} 1}}%
      \put(849,220){\makebox(0,0){\strut{} 10}}%
      \put(1434,220){\makebox(0,0){\strut{} 100}}%
      \put(2019,220){\makebox(0,0){\strut{} 1000}}%
      \put(2605,220){\makebox(0,0){\strut{} 10000}}%
      \put(3190,220){\makebox(0,0){\strut{} 100000}}%
      \put(3775,220){\makebox(0,0){\strut{} 1e+06}}%
      \put(4360,220){\makebox(0,0){\strut{} 1e+07}}%
      \put(-638,1370){\rotatebox{-270}{\makebox(0,0){\strut{}$|\vec{p}|$}}}%
      \put(2312,0){\makebox(0,0){\strut{}Iterations}}%
    }%
    \gplgaddtomacro\gplfronttext{%
      \csname LTb\endcsname%
      \put(1716,833){\makebox(0,0)[r]{\strut{}VGL(0)/DHP}}%
      \csname LTb\endcsname%
      \put(1716,613){\makebox(0,0)[r]{\strut{}VGL(1)}}%
    }%
    \gplbacktext
    \put(0,0){\includegraphics{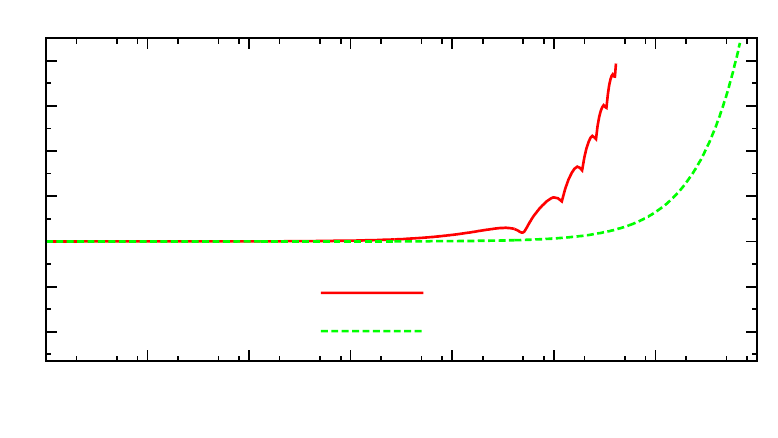}}%
    \gplfronttext
  \end{picture}%
\endgroup

%% file: plotVGLOmega1convergencet.tex
% GNUPLOT: LaTeX picture with Postscript
\begingroup
  \makeatletter
  \providecommand\color[2][]{%
    \GenericError{(gnuplot) \space\space\space\@spaces}{%
      Package color not loaded in conjunction with
      terminal option `colourtext'%
    }{See the gnuplot documentation for explanation.%
    }{Either use 'blacktext' in gnuplot or load the package
      color.sty in LaTeX.}%
    \renewcommand\color[2][]{}%
  }%
  \providecommand\includegraphics[2][]{%
    \GenericError{(gnuplot) \space\space\space\@spaces}{%
      Package graphicx or graphics not loaded%
    }{See the gnuplot documentation for explanation.%
    }{The gnuplot epslatex terminal needs graphicx.sty or graphics.sty.}%
    \renewcommand\includegraphics[2][]{}%
  }%
  \providecommand\rotatebox[2]{#2}%
  \@ifundefined{ifGPcolor}{%
    \newif\ifGPcolor
    \GPcolortrue
  }{}%
  \@ifundefined{ifGPblacktext}{%
    \newif\ifGPblacktext
    \GPblacktextfalse
  }{}%
  % define a \g@addto@macro without @ in the name:
  \let\gplgaddtomacro\g@addto@macro
  % define empty templates for all commands taking text:
  \gdef\gplbacktext{}%
  \gdef\gplfronttext{}%
  \makeatother
  \ifGPblacktext
    % no textcolor at all
    \def\colorrgb#1{}%
    \def\colorgray#1{}%
  \else
    % gray or color?
    \ifGPcolor
      \def\colorrgb#1{\color[rgb]{#1}}%
      \def\colorgray#1{\color[gray]{#1}}%
      \expandafter\def\csname LTw\endcsname{\color{white}}%
      \expandafter\def\csname LTb\endcsname{\color{black}}%
      \expandafter\def\csname LTa\endcsname{\color{black}}%
      \expandafter\def\csname LT0\endcsname{\color[rgb]{1,0,0}}%
      \expandafter\def\csname LT1\endcsname{\color[rgb]{0,1,0}}%
      \expandafter\def\csname LT2\endcsname{\color[rgb]{0,0,1}}%
      \expandafter\def\csname LT3\endcsname{\color[rgb]{1,0,1}}%
      \expandafter\def\csname LT4\endcsname{\color[rgb]{0,1,1}}%
      \expandafter\def\csname LT5\endcsname{\color[rgb]{1,1,0}}%
      \expandafter\def\csname LT6\endcsname{\color[rgb]{0,0,0}}%
      \expandafter\def\csname LT7\endcsname{\color[rgb]{1,0.3,0}}%
      \expandafter\def\csname LT8\endcsname{\color[rgb]{0.5,0.5,0.5}}%
    \else
      % gray
      \def\colorrgb#1{\color{black}}%
      \def\colorgray#1{\color[gray]{#1}}%
      \expandafter\def\csname LTw\endcsname{\color{white}}%
      \expandafter\def\csname LTb\endcsname{\color{black}}%
      \expandafter\def\csname LTa\endcsname{\color{black}}%
      \expandafter\def\csname LT0\endcsname{\color{black}}%
      \expandafter\def\csname LT1\endcsname{\color{black}}%
      \expandafter\def\csname LT2\endcsname{\color{black}}%
      \expandafter\def\csname LT3\endcsname{\color{black}}%
      \expandafter\def\csname LT4\endcsname{\color{black}}%
      \expandafter\def\csname LT5\endcsname{\color{black}}%
      \expandafter\def\csname LT6\endcsname{\color{black}}%
      \expandafter\def\csname LT7\endcsname{\color{black}}%
      \expandafter\def\csname LT8\endcsname{\color{black}}%
    \fi
  \fi
  \setlength{\unitlength}{0.0500bp}%
  \begin{picture}(4492.80,2520.00)%
    \gplgaddtomacro\gplbacktext{%
      \csname LTb\endcsname%
      \put(132,440){\makebox(0,0)[r]{\strut{} 1e-10}}%
      \put(132,647){\makebox(0,0)[r]{\strut{} 1e-09}}%
      \put(132,853){\makebox(0,0)[r]{\strut{} 1e-08}}%
      \put(132,1060){\makebox(0,0)[r]{\strut{} 1e-07}}%
      \put(132,1267){\makebox(0,0)[r]{\strut{} 1e-06}}%
      \put(132,1473){\makebox(0,0)[r]{\strut{} 1e-05}}%
      \put(132,1680){\makebox(0,0)[r]{\strut{} 0.0001}}%
      \put(132,1887){\makebox(0,0)[r]{\strut{} 0.001}}%
      \put(132,2093){\makebox(0,0)[r]{\strut{} 0.01}}%
      \put(132,2300){\makebox(0,0)[r]{\strut{} 0.1}}%
      \put(264,220){\makebox(0,0){\strut{} 1}}%
      \put(849,220){\makebox(0,0){\strut{} 10}}%
      \put(1434,220){\makebox(0,0){\strut{} 100}}%
      \put(2019,220){\makebox(0,0){\strut{} 1000}}%
      \put(2605,220){\makebox(0,0){\strut{} 10000}}%
      \put(3190,220){\makebox(0,0){\strut{} 100000}}%
      \put(3775,220){\makebox(0,0){\strut{} 1e+06}}%
      \put(4360,220){\makebox(0,0){\strut{} 1e+07}}%
      \put(-638,1370){\rotatebox{-270}{\makebox(0,0){\strut{}$|\vec{p}|$}}}%
      \put(2312,0){\makebox(0,0){\strut{}Iterations}}%
    }%
    \gplgaddtomacro\gplfronttext{%
      \csname LTb\endcsname%
      \put(2244,613){\makebox(0,0)[r]{\strut{}VGL$\Omega$(1)}}%
    }%
    \gplbacktext
    \put(0,0){\includegraphics{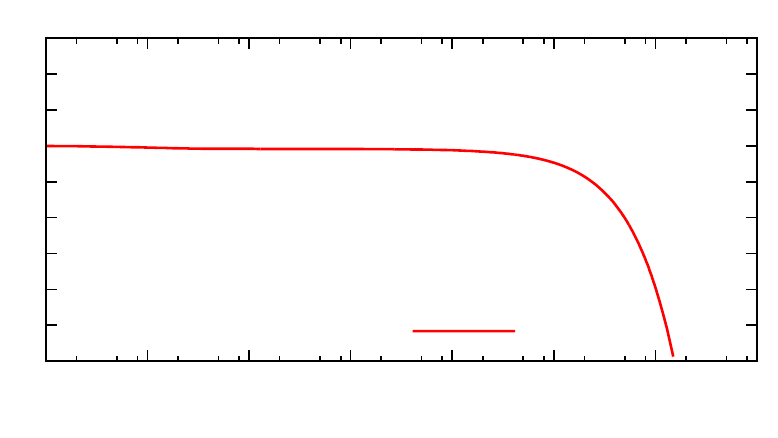}}%
    \gplfronttext
  \end{picture}%
\endgroup

%% file: plotTD0divergencet.tex
% GNUPLOT: LaTeX picture with Postscript
\begingroup
  \makeatletter
  \providecommand\color[2][]{%
    \GenericError{(gnuplot) \space\space\space\@spaces}{%
      Package color not loaded in conjunction with
      terminal option `colourtext'%
    }{See the gnuplot documentation for explanation.%
    }{Either use 'blacktext' in gnuplot or load the package
      color.sty in LaTeX.}%
    \renewcommand\color[2][]{}%
  }%
  \providecommand\includegraphics[2][]{%
    \GenericError{(gnuplot) \space\space\space\@spaces}{%
      Package graphicx or graphics not loaded%
    }{See the gnuplot documentation for explanation.%
    }{The gnuplot epslatex terminal needs graphicx.sty or graphics.sty.}%
    \renewcommand\includegraphics[2][]{}%
  }%
  \providecommand\rotatebox[2]{#2}%
  \@ifundefined{ifGPcolor}{%
    \newif\ifGPcolor
    \GPcolortrue
  }{}%
  \@ifundefined{ifGPblacktext}{%
    \newif\ifGPblacktext
    \GPblacktextfalse
  }{}%
  % define a \g@addto@macro without @ in the name:
  \let\gplgaddtomacro\g@addto@macro
  % define empty templates for all commands taking text:
  \gdef\gplbacktext{}%
  \gdef\gplfronttext{}%
  \makeatother
  \ifGPblacktext
    % no textcolor at all
    \def\colorrgb#1{}%
    \def\colorgray#1{}%
  \else
    % gray or color?
    \ifGPcolor
      \def\colorrgb#1{\color[rgb]{#1}}%
      \def\colorgray#1{\color[gray]{#1}}%
      \expandafter\def\csname LTw\endcsname{\color{white}}%
      \expandafter\def\csname LTb\endcsname{\color{black}}%
      \expandafter\def\csname LTa\endcsname{\color{black}}%
      \expandafter\def\csname LT0\endcsname{\color[rgb]{1,0,0}}%
      \expandafter\def\csname LT1\endcsname{\color[rgb]{0,1,0}}%
      \expandafter\def\csname LT2\endcsname{\color[rgb]{0,0,1}}%
      \expandafter\def\csname LT3\endcsname{\color[rgb]{1,0,1}}%
      \expandafter\def\csname LT4\endcsname{\color[rgb]{0,1,1}}%
      \expandafter\def\csname LT5\endcsname{\color[rgb]{1,1,0}}%
      \expandafter\def\csname LT6\endcsname{\color[rgb]{0,0,0}}%
      \expandafter\def\csname LT7\endcsname{\color[rgb]{1,0.3,0}}%
      \expandafter\def\csname LT8\endcsname{\color[rgb]{0.5,0.5,0.5}}%
    \else
      % gray
      \def\colorrgb#1{\color{black}}%
      \def\colorgray#1{\color[gray]{#1}}%
      \expandafter\def\csname LTw\endcsname{\color{white}}%
      \expandafter\def\csname LTb\endcsname{\color{black}}%
      \expandafter\def\csname LTa\endcsname{\color{black}}%
      \expandafter\def\csname LT0\endcsname{\color{black}}%
      \expandafter\def\csname LT1\endcsname{\color{black}}%
      \expandafter\def\csname LT2\endcsname{\color{black}}%
      \expandafter\def\csname LT3\endcsname{\color{black}}%
      \expandafter\def\csname LT4\endcsname{\color{black}}%
      \expandafter\def\csname LT5\endcsname{\color{black}}%
      \expandafter\def\csname LT6\endcsname{\color{black}}%
      \expandafter\def\csname LT7\endcsname{\color{black}}%
      \expandafter\def\csname LT8\endcsname{\color{black}}%
    \fi
  \fi
  \setlength{\unitlength}{0.0500bp}%
  \begin{picture}(4492.80,2520.00)%
    \gplgaddtomacro\gplbacktext{%
      \csname LTb\endcsname%
      \put(132,440){\makebox(0,0)[r]{\strut{} 0}}%
      \put(132,812){\makebox(0,0)[r]{\strut{} 0.002}}%
      \put(132,1184){\makebox(0,0)[r]{\strut{} 0.004}}%
      \put(132,1556){\makebox(0,0)[r]{\strut{} 0.006}}%
      \put(132,1928){\makebox(0,0)[r]{\strut{} 0.008}}%
      \put(132,2300){\makebox(0,0)[r]{\strut{} 0.01}}%
      \put(264,220){\makebox(0,0){\strut{} 0}}%
      \put(947,220){\makebox(0,0){\strut{} 5e+09}}%
      \put(1629,220){\makebox(0,0){\strut{} 1e+10}}%
      \put(2312,220){\makebox(0,0){\strut{} 1.5e+10}}%
      \put(2995,220){\makebox(0,0){\strut{} 2e+10}}%
      \put(3677,220){\makebox(0,0){\strut{} 2.5e+10}}%
      \put(4360,220){\makebox(0,0){\strut{} 3e+10}}%
      \put(-506,1370){\rotatebox{-270}{\makebox(0,0){\strut{}$|\vec{p}|$}}}%
      \put(2312,0){\makebox(0,0){\strut{}Iterations}}%
      \put(2312,2498){\makebox(0,0){\strut{}Divergence of TD(0)/Sarsa(0)}}%
    }%
    \gplgaddtomacro\gplfronttext{%
    }%
    \gplbacktext
    \put(0,0){\includegraphics{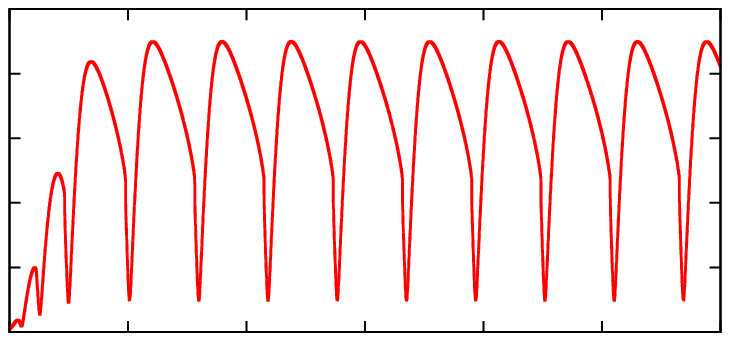}}%
    \gplfronttext
  \end{picture}%
\endgroup

%% file: plotTD0divergencePhaseSpacet.tex
% GNUPLOT: LaTeX picture with Postscript
\begingroup
  \makeatletter
  \providecommand\color[2][]{%
    \GenericError{(gnuplot) \space\space\space\@spaces}{%
      Package color not loaded in conjunction with
      terminal option `colourtext'%
    }{See the gnuplot documentation for explanation.%
    }{Either use 'blacktext' in gnuplot or load the package
      color.sty in LaTeX.}%
    \renewcommand\color[2][]{}%
  }%
  \providecommand\includegraphics[2][]{%
    \GenericError{(gnuplot) \space\space\space\@spaces}{%
      Package graphicx or graphics not loaded%
    }{See the gnuplot documentation for explanation.%
    }{The gnuplot epslatex terminal needs graphicx.sty or graphics.sty.}%
    \renewcommand\includegraphics[2][]{}%
  }%
  \providecommand\rotatebox[2]{#2}%
  \@ifundefined{ifGPcolor}{%
    \newif\ifGPcolor
    \GPcolortrue
  }{}%
  \@ifundefined{ifGPblacktext}{%
    \newif\ifGPblacktext
    \GPblacktextfalse
  }{}%
  % define a \g@addto@macro without @ in the name:
  \let\gplgaddtomacro\g@addto@macro
  % define empty templates for all commands taking text:
  \gdef\gplbacktext{}%
  \gdef\gplfronttext{}%
  \makeatother
  \ifGPblacktext
    % no textcolor at all
    \def\colorrgb#1{}%
    \def\colorgray#1{}%
  \else
    % gray or color?
    \ifGPcolor
      \def\colorrgb#1{\color[rgb]{#1}}%
      \def\colorgray#1{\color[gray]{#1}}%
      \expandafter\def\csname LTw\endcsname{\color{white}}%
      \expandafter\def\csname LTb\endcsname{\color{black}}%
      \expandafter\def\csname LTa\endcsname{\color{black}}%
      \expandafter\def\csname LT0\endcsname{\color[rgb]{1,0,0}}%
      \expandafter\def\csname LT1\endcsname{\color[rgb]{0,1,0}}%
      \expandafter\def\csname LT2\endcsname{\color[rgb]{0,0,1}}%
      \expandafter\def\csname LT3\endcsname{\color[rgb]{1,0,1}}%
      \expandafter\def\csname LT4\endcsname{\color[rgb]{0,1,1}}%
      \expandafter\def\csname LT5\endcsname{\color[rgb]{1,1,0}}%
      \expandafter\def\csname LT6\endcsname{\color[rgb]{0,0,0}}%
      \expandafter\def\csname LT7\endcsname{\color[rgb]{1,0.3,0}}%
      \expandafter\def\csname LT8\endcsname{\color[rgb]{0.5,0.5,0.5}}%
    \else
      % gray
      \def\colorrgb#1{\color{black}}%
      \def\colorgray#1{\color[gray]{#1}}%
      \expandafter\def\csname LTw\endcsname{\color{white}}%
      \expandafter\def\csname LTb\endcsname{\color{black}}%
      \expandafter\def\csname LTa\endcsname{\color{black}}%
      \expandafter\def\csname LT0\endcsname{\color{black}}%
      \expandafter\def\csname LT1\endcsname{\color{black}}%
      \expandafter\def\csname LT2\endcsname{\color{black}}%
      \expandafter\def\csname LT3\endcsname{\color{black}}%
      \expandafter\def\csname LT4\endcsname{\color{black}}%
      \expandafter\def\csname LT5\endcsname{\color{black}}%
      \expandafter\def\csname LT6\endcsname{\color{black}}%
      \expandafter\def\csname LT7\endcsname{\color{black}}%
      \expandafter\def\csname LT8\endcsname{\color{black}}%
    \fi
  \fi
  \setlength{\unitlength}{0.0500bp}%
  \begin{picture}(4492.80,2520.00)%
    \gplgaddtomacro\gplbacktext{%
      \csname LTb\endcsname%
      \put(132,440){\makebox(0,0)[r]{\strut{}-0.002}}%
      \put(132,673){\makebox(0,0)[r]{\strut{}-0.0015}}%
      \put(132,905){\makebox(0,0)[r]{\strut{}-0.001}}%
      \put(132,1138){\makebox(0,0)[r]{\strut{}-0.0005}}%
      \put(132,1370){\makebox(0,0)[r]{\strut{} 0}}%
      \put(132,1603){\makebox(0,0)[r]{\strut{} 0.0005}}%
      \put(132,1835){\makebox(0,0)[r]{\strut{} 0.001}}%
      \put(132,2068){\makebox(0,0)[r]{\strut{} 0.0015}}%
      \put(132,2300){\makebox(0,0)[r]{\strut{} 0.002}}%
      \put(636,220){\makebox(0,0){\strut{} 0}}%
      \put(1381,220){\makebox(0,0){\strut{} 0.002}}%
      \put(2126,220){\makebox(0,0){\strut{} 0.004}}%
      \put(2871,220){\makebox(0,0){\strut{} 0.006}}%
      \put(3615,220){\makebox(0,0){\strut{} 0.008}}%
      \put(4360,220){\makebox(0,0){\strut{} 0.01}}%
      \put(-638,1370){\rotatebox{-270}{\makebox(0,0){\strut{}$p_2$}}}%
      \put(2312,0){\makebox(0,0){\strut{}$p_1$}}%
      \put(2312,2498){\makebox(0,0){\strut{}Phase Space for Divergence of TD(0)/Sarsa(0)}}%
    }%
    \gplgaddtomacro\gplfronttext{%
    }%
    \gplbacktext
    \put(0,0){\includegraphics{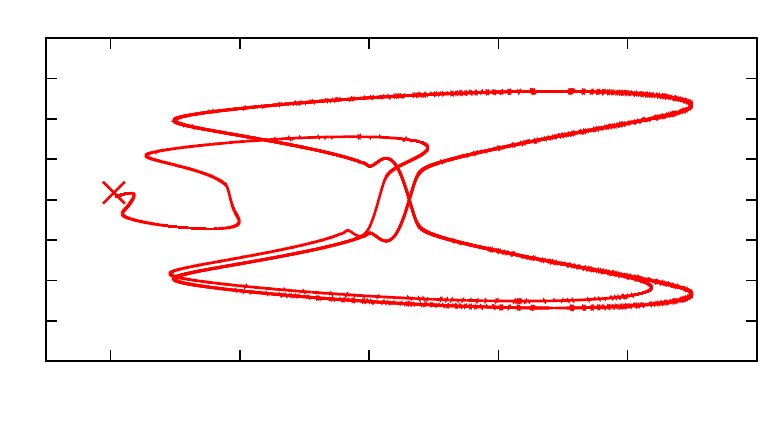}}%
    \gplfronttext
  \end{picture}%
\endgroup

%% file: plotTD1divergencet.tex
% GNUPLOT: LaTeX picture with Postscript
\begingroup
  \makeatletter
  \providecommand\color[2][]{%
    \GenericError{(gnuplot) \space\space\space\@spaces}{%
      Package color not loaded in conjunction with
      terminal option `colourtext'%
    }{See the gnuplot documentation for explanation.%
    }{Either use 'blacktext' in gnuplot or load the package
      color.sty in LaTeX.}%
    \renewcommand\color[2][]{}%
  }%
  \providecommand\includegraphics[2][]{%
    \GenericError{(gnuplot) \space\space\space\@spaces}{%
      Package graphicx or graphics not loaded%
    }{See the gnuplot documentation for explanation.%
    }{The gnuplot epslatex terminal needs graphicx.sty or graphics.sty.}%
    \renewcommand\includegraphics[2][]{}%
  }%
  \providecommand\rotatebox[2]{#2}%
  \@ifundefined{ifGPcolor}{%
    \newif\ifGPcolor
    \GPcolortrue
  }{}%
  \@ifundefined{ifGPblacktext}{%
    \newif\ifGPblacktext
    \GPblacktextfalse
  }{}%
  % define a \g@addto@macro without @ in the name:
  \let\gplgaddtomacro\g@addto@macro
  % define empty templates for all commands taking text:
  \gdef\gplbacktext{}%
  \gdef\gplfronttext{}%
  \makeatother
  \ifGPblacktext
    % no textcolor at all
    \def\colorrgb#1{}%
    \def\colorgray#1{}%
  \else
    % gray or color?
    \ifGPcolor
      \def\colorrgb#1{\color[rgb]{#1}}%
      \def\colorgray#1{\color[gray]{#1}}%
      \expandafter\def\csname LTw\endcsname{\color{white}}%
      \expandafter\def\csname LTb\endcsname{\color{black}}%
      \expandafter\def\csname LTa\endcsname{\color{black}}%
      \expandafter\def\csname LT0\endcsname{\color[rgb]{1,0,0}}%
      \expandafter\def\csname LT1\endcsname{\color[rgb]{0,1,0}}%
      \expandafter\def\csname LT2\endcsname{\color[rgb]{0,0,1}}%
      \expandafter\def\csname LT3\endcsname{\color[rgb]{1,0,1}}%
      \expandafter\def\csname LT4\endcsname{\color[rgb]{0,1,1}}%
      \expandafter\def\csname LT5\endcsname{\color[rgb]{1,1,0}}%
      \expandafter\def\csname LT6\endcsname{\color[rgb]{0,0,0}}%
      \expandafter\def\csname LT7\endcsname{\color[rgb]{1,0.3,0}}%
      \expandafter\def\csname LT8\endcsname{\color[rgb]{0.5,0.5,0.5}}%
    \else
      % gray
      \def\colorrgb#1{\color{black}}%
      \def\colorgray#1{\color[gray]{#1}}%
      \expandafter\def\csname LTw\endcsname{\color{white}}%
      \expandafter\def\csname LTb\endcsname{\color{black}}%
      \expandafter\def\csname LTa\endcsname{\color{black}}%
      \expandafter\def\csname LT0\endcsname{\color{black}}%
      \expandafter\def\csname LT1\endcsname{\color{black}}%
      \expandafter\def\csname LT2\endcsname{\color{black}}%
      \expandafter\def\csname LT3\endcsname{\color{black}}%
      \expandafter\def\csname LT4\endcsname{\color{black}}%
      \expandafter\def\csname LT5\endcsname{\color{black}}%
      \expandafter\def\csname LT6\endcsname{\color{black}}%
      \expandafter\def\csname LT7\endcsname{\color{black}}%
      \expandafter\def\csname LT8\endcsname{\color{black}}%
    \fi
  \fi
  \setlength{\unitlength}{0.0500bp}%
  \begin{picture}(4492.80,2520.00)%
    \gplgaddtomacro\gplbacktext{%
      \csname LTb\endcsname%
      \put(132,680){\makebox(0,0)[r]{\strut{} 1e-08}}%
      \put(132,1049){\makebox(0,0)[r]{\strut{} 1e-06}}%
      \put(132,1418){\makebox(0,0)[r]{\strut{} 0.0001}}%
      \put(132,1787){\makebox(0,0)[r]{\strut{} 0.01}}%
      \put(132,2156){\makebox(0,0)[r]{\strut{} 1}}%
      \put(264,220){\makebox(0,0){\strut{} 1}}%
      \put(1009,220){\makebox(0,0){\strut{} 100}}%
      \put(1753,220){\makebox(0,0){\strut{} 10000}}%
      \put(2498,220){\makebox(0,0){\strut{} 1e+06}}%
      \put(3243,220){\makebox(0,0){\strut{} 1e+08}}%
      \put(3988,220){\makebox(0,0){\strut{} 1e+10}}%
      \put(-638,1370){\rotatebox{-270}{\makebox(0,0){\strut{}$|\vec{p}|$}}}%
      \put(2312,0){\makebox(0,0){\strut{}Iterations}}%
    }%
    \gplgaddtomacro\gplfronttext{%
      \csname LTb\endcsname%
      \put(2244,613){\makebox(0,0)[r]{\strut{}TD(1)/Sarsa(1)}}%
    }%
    \gplbacktext
    \put(0,0){\includegraphics{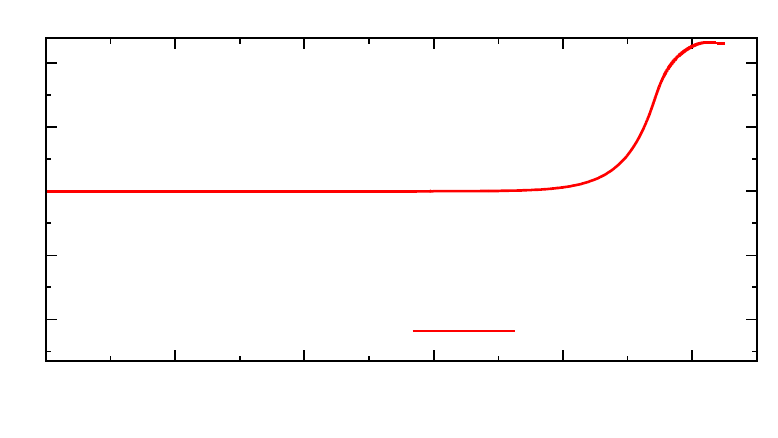}}%
    \gplfronttext
  \end{picture}%
\endgroup